\documentclass{article} 
\usepackage{iclr2025_conference, times}
\usepackage{graphicx}

\usepackage{amsmath,amsfonts,bm}

\def\eqref#1{equation~\ref{#1}}

\def\1{\bm{1}}

\DeclareMathAlphabet{\mathsfit}{\encodingdefault}{\sfdefault}{m}{sl}
\SetMathAlphabet{\mathsfit}{bold}{\encodingdefault}{\sfdefault}{bx}{n}

\usepackage{graphicx}
\usepackage{hyperref}
\usepackage{url}
\usepackage{booktabs}
\title{Vision Language Models See What You Want but not What You See}
\iclrfinalcopy
\author{%
  Qingying Gao\textsuperscript{1}, Yijiang Li\textsuperscript{2}, Haiyun Lyu\textsuperscript{3}, Haoran Sun\textsuperscript{1}, Dezhi Luo\textsuperscript{4}, Hokin Deng\textsuperscript{5}\thanks{Correspondence to Yijiang Li (yijiangli@ucsd.edu), Dezhi Luo (ihzedoul@umich.edu), Hokin Deng (hokind@andrew.cmu.edu).}
  \\
  \textsuperscript{1}Johns Hopkins University, \textsuperscript{2}University of California San Diego,
  \\ \textsuperscript{3}University of North Carolina at Chapel Hill ,\textsuperscript{4}University of Michigan,
  \\ \textsuperscript{5}Carnegie Mellon University
  \\ All authors are affiliated with \href{https://growing-ai-like-a-child.github.io/}{GrowAI}
}

\begin{document}

\maketitle

\begin{abstract}
Knowing others' intentions and taking others' perspectives are two core components of human intelligence that are considered to be instantiations of theory-of-mind. Infiltrating machines with these abilities is an important step towards building human-level artificial intelligence. Here, to investigate intentionality understanding and level-2 perspective-taking in Vision Language Models (VLMs), we constructed the IntentBench and PerspectBench, which together contains over 300 cognitive experiments grounded in real-world scenarios and classic cognitive tasks. We found VLMs achieving high performance on intentionality understanding but low performance on level-2 perspective-taking. This suggests a potential dissociation between simulation-based and theory-based theory-of-mind abilities in VLMs, highlighting the concern that they are not capable of using model-based reasoning to infer others' mental states.

\textbf{Keywords:} 
vision language models; perspective-taking; intentionality understanding; theory-of-mind; cognitive AI
\end{abstract}

\section{Introduction}

Intentionality is the capacity of the mind to be directed toward, represent, or stand for objects, properties, or states of affairs for further executable actions \citep{anscombe1956intention}. To say one could understand intentionality is to say one has the capacity to comprehend the mental content for action in another mind \citep{premack1978chimpanzee,rosenthal1991nature}. This capacity has been seen as a key distinction between humans and machines \citep{searle1980minds}. It is argued that despite well manipulation of language symbols, machines cannot understand intentional meanings of others because it lacks theory-of-mind (ToM), the kind of abilities that allows one to infer the mental content of others \citep{premack1978chimpanzee,dennett1987intentional}. Nevertheless, several recent studies have showed that large language models (LLMs) and their supporting vision language models (VLMs) exhibit ToM abilities \citep{Kosinski2023ToM, li2025egoprivacy, Strachan2024ToM, shi2024muma}, thus calling for updated examinations of the nature of ToM and the potential for current and future artificial intelligence to possess such abilities.

We believe an important approach to said inquiry is examining the extent to which different ToM abilities necessitate model-based reasoning. Specifically, a distinction can be drawn between ToM abilities based on simulation-theory and theory-theory. The former involves the construction of an internal model of self-other relations to reason about the mental states of others, whereas the latter requires only the use of theoretical knowledge regarding the relations between mind and behavior \citep{gopnik1992child, frith2005theory, shanton2010simulation}. Whether current artificial intelligence systems possess internal models that are available for reasoning remains a key debate, with several influential accounts questioning the existence of model-based reasoning among LLMs \citep{hao2023reasoning, mitchell2023debate, yildirim2024task, li2022more, goddu2024llms}. If this is indeed the case, then evidences regarding the possession of ToM abilities in VLMs above would imply that ToM abilities do not require mental simulation, and that mental simulation is not within the foundational capabilities of ToM systems. 

We tested this critical prediction by assessing VLMs' ability to perform intentionality understanding and level-2 perspective-taking. ToM is commonly understood to be grounded in perspective-taking, a series of multi-level abilities that involves the cognitively undertaking of the perspective of another \citep{Barnes-Holmes2004perspective}. Level-1 perspective-taking refers to the acknowledgement that different people can see different things, whereas level-2 perspective-taking involves the understanding of how another person may see the same thing differently. While level-1 perspective-taking emerges in humans as early as 2 years old, much older children are found to struggle with level-2 perspective-taking \citep{piaget1977development}. This is likely because, despite its relatively low level in the perspective-taking hierarchy, this ability requires model-based reasoning, exemplified in the visual domain as inferences based on mental rotation \citep{lehmann2019relationship, gunia2021brain}. On the other hand, as another ability at the core of human Theory-of-Mind, while intentionality understanding involves high-level cognition and abstract reasoning, it is unclear whether this complex ability necessitate mental simulations\citep{wellman1992child, frith2006neural, Apperly2010mindreaders, Kilner2011action, Bianco2024action}. Assessing level-2 perspective-taking and intentionality understanding in VLMs could thus provide insights into not only VLMs' abilities within these two ToM domains but also whether the comparative performance between them follows that of humans. However, these areas remain largely unexplored in the current literature, and in intelligent systems beyond human beings.

To address this critical gap in the literature, we leveraged and further instantiated the IntentBench and PerspectBench of \textbf{CoreCognition} benchmark \citep{li2024core}, two targeted dataset designed to systematically evaluate the capabilities of current VLMs in intentionality understanding and level-2 perspective-taking, respectively.

\section{Methods}

\subsection{Dataset} 

PerspectBench consists of 32 multi-image and 209 single-image format experiments based on classic cognitive tasks. IntentBench consists of 100 single-image format experiments based on real-world ambiguous social scenarios.

\subsection{Experiment Design}

\subsubsection{Level-2 Perspective-taking} In Piagetian developmental psychology, the acquisition of level-2 perspective-taking ability marks a milestone of human cognitive development as it indicates the elimination of egocentrism -- the inability to consider perspectives other than one's own \citep{piaget1977development}. The Three Mountain Task invented by Piaget is widely used in developmental psychology laboratories as the gold standard for testing level-1 and level-2 perspective-taking abilities in children \citep{piaget1956conception, johnson1975cooperativeness, ford1979construct, lamm2007neural}. In a standard Three Mountain Task assessment, a child is instructed to position oneself in front of a model featuring three mountains. These mountains vary in size and are distinguished by unique characteristics: one is covered in snow, another has a red cross at its peak, and the third is topped with a hut. The child is then asked to perform a complete 360-degree examination of the model. Subsequently, another individual is introduced and takes a different vantage point to observe the model. The child is presented with several photographs that showcase various viewpoints of the model and is tasked with identifying which photograph accurately represents what the other person sees. At around four years of age, children typically select the photograph that matches their own perspective. By six years old, they begin to acknowledge viewpoints that differ from their own, and by the ages of seven to eight, they are generally able to reliably identify the perspective of another individual \citep{piaget1969psychology}. To test level-2 perspective-taking in VLMs, we adapted the Three Mountain task into formats that are suitable for benchmarks with minimal confounding details while preserving real-life spatiality. In particular, we used groups of 3-4 commonly-seen elastic cans organized into different spatial patterns to mimic the mountain model. Like in the original task, we used a doll placed to face the organization from different angels as the object of perspective-taking.

\begin{figure*}[t]
\centering
\includegraphics[width=0.9\textwidth]{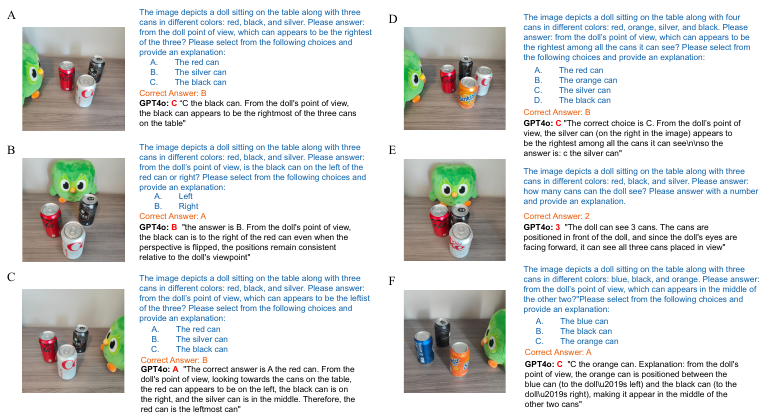}
\caption{\textbf{Example Experiments and Model Performances on PerspectBench}}
\label{fig1}
\end{figure*}

\subsubsection{Intentionality Understanding} Intentionality understanding is believed to be grounded by rudimentary theory-of-mind abilities \citep{wellman1992child, frith2006neural, Apperly2010mindreaders}. In developmental psychology, a critical subset of intentionality understanding experiments involves tests of action understanding \citep{searle1979intentionality}. Several computational hypotheses are proposed on how one could understand other people's actions: for example, action understanding could be computationally modeled as pure inference \citep{gweon201116, gweon2021inferential}, as mental action simulation \citep{brass2007investigating}, or as inverse planning \citep{baker2009action}. Typically, cartoon stimuli built via physic simulation engine are used frequently in action understanding in developmental psychology \citep{liu2017ten, shu2021agent}. These stimuli are incorporated into IntentBench. However, a common critique of cognitive psychology tasks is that they lack realism and have limited applicability to real-world situations \citep{gomez2014big}. Drawing inspiration from COIG-CQIA and its Ruozhiba dataset, many real-world ambiguous scenarios are incorporated into IntentBench for explicitly testing intentionality understanding in ethological conditions \citep{bai2024coig}.

\begin{figure*}[t]
\centering
\includegraphics[width=0.9\textwidth]{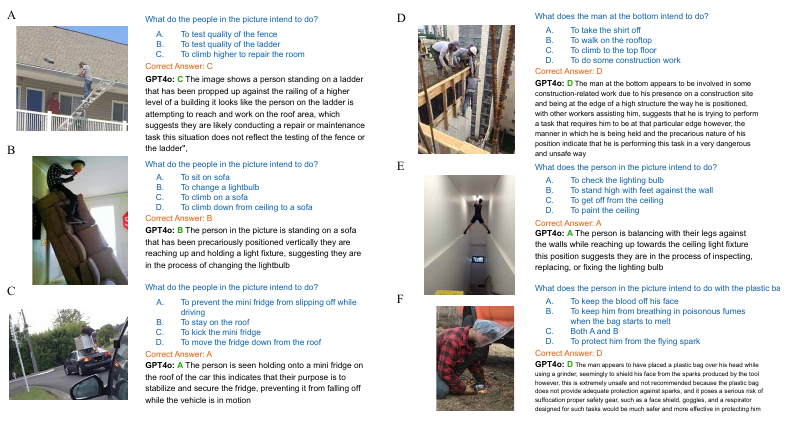}
\caption{\textbf{Example Experiments and Model Performances on IntentBench}}
\label{fig1}
\end{figure*}

\subsection{Examined Vision Language Models}

Recent advances in multi-modal learning have been largely driven by the unified modeling of visual and textual modalities using transformers \citep{li2019visualbert, xu2023bridgetower, tan2019lxmert, alayrac2022flamingo, radford2021learning}. With the rise of large language models (LLMs), state-of-the-art (SOTA) multi-modal LLMs (MLLMs) \citep{liu2024visual, li2023blip2} increasingly build on open-source LLM backbones \citep{touvron2023llama, peng2023instruction, jiang2023mistral}, aligning visual features to the LLM embedding space \citep{li2023blip, fu2023mme, wu2024v, xu2024llava, shao2024visual, li2022more, li2025egoprivacy, brown2020language, achiam2023gpt, bai2023qwen, jaech2024openai, zhang2025unified, zhang2024pixels}. These models have progressively achieved competitive results in complex tasks requiring high-level perception and reasoning \citep{li2024seed, team2023gemini, fu2023mme, openai2023gpt4}, including spatial reasoning \citep{chen2024spatialvlm, cai2024spatialbot}, character recognition \citep{mori1999optical}, scene understanding \citep{cordts2016cityscapes, wang2023consistent, li2023diverse, chen2017deeplab}, action recognition \citep{jhuang2013towards, herath2017going}, and action prediction \citep{lan2014hierarchical, kong2022human}, in some cases approaching human-level performance.

Building on these advancements, we conduct a systematic evaluation of vision–language models (VLMs) spanning three distinct origins and capacity scales. To ensure comparability, all models are assessed on a zero-shot image–text reasoning task in a generative setting. The complete list of evaluated models, along with curated model size statistics, is presented in the results section (Figure~\ref{fig3}). For analysis, we group the models into the following categories:

\begin{enumerate}
    \item \textbf{Open-source VLMs with Multi-Image Reasoning}:  
    Includes models with different sizes and other variants such as \texttt{CogVLM} Series \citep{hong2024cogvlm2}, \texttt{Qwen} series(Qwen-VL \citep{Qwen-VL}, Qwen-2 \citep{Qwen2VL}), and \texttt{Blip2} \citep{li2023blip2}, LLaVA-Next \citep{liu2024llavanext} , which are capable of reasoning over interleaved multiple images and texts.
    \item \textbf{Closed-source VLMs with Multi-Image Reasoning}:  
    Includes proprietary models such as GPT series \citep{gpt4o} ( \texttt{GPT-4v}, \texttt{GPT-4-turbo}, \texttt{GPT-4o-mini}), Gemini Series \citep{gemini}, and Claude Series \citep{claude}. These models also support reasoning across interleaved images and texts,
    \item     \textbf{Open-source VLMs with single-Image Reasoning}:  
   Includes models designed to process a single image alongside continuous text.  InstructBlip Series \citep{instructblip}, LLaVA Series \citep{liu2023improvedllava} \citep{liu2023llava} 
\end{enumerate}

In total, we processed 37 models for evaluation. All the model performances in intentionality understanding and perspective-taking, together with human baseline performances, are presented here (Figure \ref{fig1}). In order to analyze the reasoning abilities of VLMs, we ask the models to explain their answers after they have given the answers.

\subsection{Human Baseline}

We recruited a total of 22 participants, all of whom were college students proficient in English. Participants were instructed to skip any question that was ambiguously phrased or too complex to answer within 90 seconds. A question was marked as failed if the participant did not provide an answer. For each question, at least 80\% of participants needed to answer correctly; otherwise, we modified the question, and new annotators completed the revised version. The human baseline result for each question was normalized based on the number of participants who provided an answer.

\section{Results}

\subsection{Overall Performance}

\begin{figure*}[t]
\centering
\includegraphics[width=0.9\textwidth]{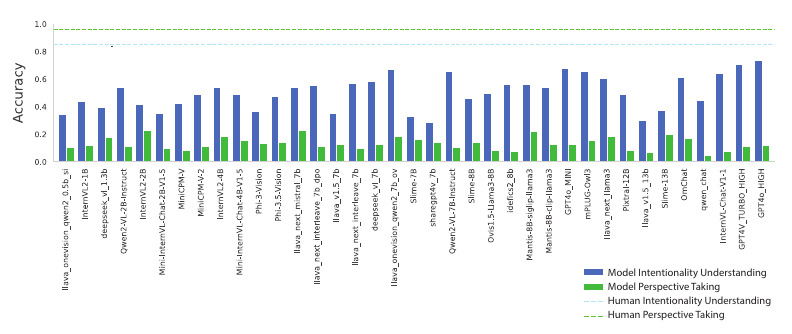}
\caption{\textbf{VLMs' Performance on IntentBench and PerspectBench As Compared to Human Baseline}}
\label{fig3}
\end{figure*}

\begin{figure}[t]
\centering
\includegraphics[width=0.9\columnwidth]{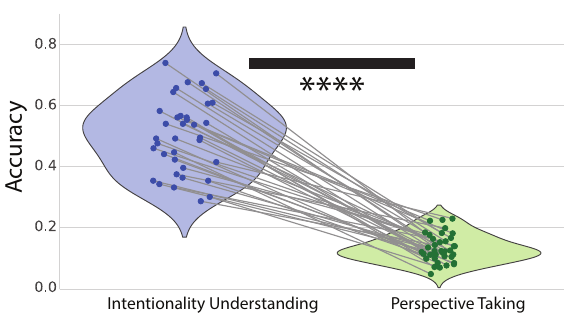}
\caption{\textbf{VLMs perform significantly better in intentionality understanding compared to perspective-taking.} Paired samples \textit{t}-test: \( p = 2.62 \times 10^{-19} \), \( t = 17.651 \) .}
\label{fig4}
\end{figure}

Our findings revealed a clear dissociation between model performance in intentionality understanding and perspective-taking. Specifically, all evaluated models demonstrate significantly stronger performance on IntentBench compared to PerspectBench (Figures \ref{fig3} and \ref{fig4}). This discrepancy becomes even more striking when compared to chance performance: while all models perform above chance performance (approximately 25.00\%) on intentionality understanding tasks, not a single model exceeds chance performance (approximately 29.03\%) on perspective-taking tasks. Notably, while some of the highest-performing models on IntentBench, such as GPT-4o, achieve near-human accuracy in intentionality comprehension, their performance on PerspectBench lags behind that of the majority of the assessed models (Figure \ref{fig4}). This gap underscores a fundamental limitation in current models, suggesting that perspective-taking might involve distinct cognitive mechanisms that are not yet fully captured by existing architectures. To quantify this disparity, we conducted a paired samples t-test on the accuracy scores of these models across the two datasets. The analysis revealed a highly significant difference in performance between the two tasks, with a t-statistic of \( t = 17.651 \) and a p-value of \( p = 2.62 \times 10^{-19} \) (Figure \ref{fig4}). This result provides strong statistical evidence that VLMs exhibit a systematic performance discrepancy, excelling at intentionality understanding while continuing to struggle with perspective-taking. These findings highlight an important challenge for the development of AI systems capable of robust social reasoning and theory of mind.

\subsection{Relationship Between Model Performance and Model Size}

\begin{figure*}[t]
\centering
\includegraphics[width=0.9\textwidth]{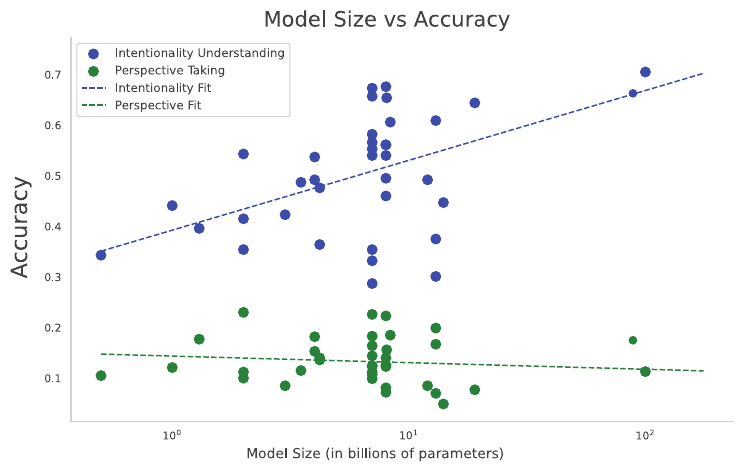}
\caption{\textbf{Differential performance changes in intentionality understanding and perspective-taking in VLMs as their model sizes increase.} Intentionality understanding: \( y = 0.0599 x + 0.3925, r^2 = 0.2797\); perspective-taking: \( y = -0.0057 x + 0.1437 , r^2 = 0.0176 \)).}
\label{fig5}
\end{figure*}

A common assumption in machine learning is that expanding a model’s scale, as measured by the number of parameters, results in systematic enhancements in its reasoning abilities \citep{sutton2019bitter, kaplan2020scaling}. We examined the degree to which this principle, known as the scaling law hypothesis, holds for the two evaluated cognitive abilities. We observed distinct trends in how intentionality understanding and perspective-taking evolve as VLMs scale in size (Figure \ref{fig5}). While larger models tend to improve in intentionality understanding, their performance in perspective-taking remains largely stagnant—or even declines slightly. This divergence raises important questions about the underlying mechanisms driving these cognitive abilities in AI models and their relationship to model scaling.

To quantitatively assess the impact of model size on performance, we conducted a linear regression analysis of accuracy scores against model size across our 37 VLMs. For intentionality understanding, the coefficient of determination is \( R^2 = 0.2797 \), with a slope of \( 0.0599 \) and an offset of \( 0.3925 \).  This positive slope indicates that as model size increases, intentionality understanding improves, which aligns with the scaling law hypothesis—the expectation that larger models generally exhibit better performance on cognitive tasks. The results for perspective-taking reveal a strikingly different trend. The coefficient of determination is much lower at \( R^2 = 0.0176 \), with a slope of \( -0.0057 \) and an offset of \( 0.1437 \). The negative slope suggests that, contrary to expectations, larger models do not show significant improvements in perspective-taking ability. In fact, their performance remains largely unchanged or even slightly decreases as they scale. This finding directly contradicts the scaling law hypothesis and suggests that perspective-taking may require fundamentally different cognitive processes that are not naturally enhanced through increased model size alone. This dissociation between intentionality understanding and perspective-taking as models scale highlights a potential limitation in current VLMs. While intentionality understanding appears to benefit from increased parameters and training data, perspective-taking does not seem to follow the same trajectory.

\subsection{Intercorrelation Between Abilities}

We further asked: Are the abilities of intentionality understanding and perspective-taking correlated in VLMs? To investigate this, we computed both Pearson and Spearman correlations between the two measures. The results indicate that there is essentially no relationship between them. The Pearson correlation coefficient is \( 0.0252 \) with \(  p = 0.882  \) and the Spearman correlation coefficient is  \( 0.0115  \) with \( p = 0.946 \). Both results suggest a lack of statistical significance, reinforcing the idea that these two cognitive abilities are largely independent within our assessment of VLMs.

\section{Discussions}

In the present work, we assessed VLMs' ability of intentionality understanding and (level-2) perspective-taking. Our results indicated that VLMs appear to be proficient in intentionality understanding while performing significantly worse in perspective-taking. 

At the higher levels of the ToM hierarchy, understanding others' intentions requires complex cognitive reasoning about abstract mental states, such as beliefs and values. While this process is cognitively demanding, previous studies suggest that intention understanding may not require explicit perspective-taking but can instead rely on contextual cues, exploiting correlations between environmental features and depicted actions through associative learning \citep{Kilner2011action, Bianco2024action}. However, intentionality understanding develops much later than level-2 perspective-taking \citep{Moll2011look}. This developmental gap has made it difficult to directly assess their functional (in)dependency using human participants. By examining these abilities in VLMs, our study likely represents the first direct investigation into their theoretical relationship, highlighting the potential of AI as a theoretical tool for cognitive science \citep{van2024reclaiming}.

Furthermore, the observed relationship between model performance on these two abilities and model size carries significant implications for VLM development. The stark contrast in how intentionality understanding and perspective-taking evolve with increasing model size suggests a fundamental difference in the scalability of these abilities within the current architectural paradigm of VLMs. The steady improvement in intentionality understanding, from near-chance performance in smaller models to near-human performance in the largest models, indicates that the attention-based architectures underpinning these models are well-suited for this ability. This suggests that scaling up model parameters is an effective and reliable approach for enhancing intentionality comprehension. In contrast, the persistent failure of all models to exceed chance-level performance on perspective-taking tasks, with larger models performing no better—and even slightly worse—than smaller ones, suggests that this ability may depend on cognitive mechanisms that the current architectures do not support. This finding implies that perspective-taking is not a scalable ability under the current model paradigm and may require fundamental architectural innovations. 

As discussed above, level-2 perspective-taking is believed to require model-based reasoning—the ability to construct an internal model of the world to support mental operations, particularly in the visual domain \citep{johnson1983mental, lehmann2019relationship, gunia2021brain}. Our findings reinforce concerns that this hallmark ability of human intelligence might remain absent in VLMs \citep{mitchell2023debate, goddu2024llms}. Moreover, given the lack of scalability observed in level-2 perspective-taking, it is possible that model-based reasoning is fundamentally unacquirable within the current architectural framework of VLMs. This raises important questions about whether alternative approaches—such as explicit world modeling architectures—are necessary to enable model-based reasoning in artificial systems \citep{lecun2022path, goddu2024llms}

One potential concern regarding the experimental paradigms in this study is the discrepancy between the setups used in IntentBench and PerspectBench: the former is based on complex, real-world scenarios, while the latter adapts controlled laboratory designs. This raises the question of whether biases inherent in these differences contribute to the observed performance gap between the two abilities. For instance, images similar to those in IntentBench may be more frequently represented in the training data, whereas models may be less familiar with minimal-context experimental setups like those in PerspectBench. Another concern is whether PerspectBench's design introduces visual confounds that could explain the poor performance. Given recent findings that VLMs struggle with basic visual recognition tasks involving simple shapes and patterns, it is possible that they fail the Three Mountain Task adaptation not due to cognitive limitations but because of visual recognition difficulties \citep{rahmanzadehgervi2024vision}. 

However, this concern is largely mitigated by the performance patterns observed in PerspectBench. If VLMs primarily struggled with image processing, their scores would cluster around chance level. Instead, all models performed significantly below chance, suggesting a systematic cognitive bias rather than a failure to interpret visual information. This pattern closely mirrors that of children struggling with level-2 perspective-taking tasks, who consistently report what they see from their own perspective rather than considering how others perceive the situation—what Piaget referred to as "egocentrism" \citep{piaget1956conception, Moll2011look, zhao2016nine}. Simply put, our results indicate that VLMs are egocentric—but not blind.

\section{Conclusion}

Overall, our study represents the first attempt to evaluate VLMs' performance in intentionality understanding and perspective-taking. Our findings suggest that while current VLMs can infer the intentions behind others’ actions, they struggle with level-2 perspective-taking. On one hand, this supports the hypothesis that intentionality understanding may not require mental simulation but could instead rely entirely on knowledge-based reasoning. On the other hand, it raises concerns that VLMs lack internal models for reasoning or, at the very least, are unable to leverage them effectively for perspective-taking. This concern is particularly significant given that intentionality understanding improves with model scale, whereas perspective-taking does not. Further research is needed to investigate these findings, as they appear to be crucial for understanding the nature of ToM abilities and their artificial implementations. Exploring the underlying mechanisms behind this dissociation may provide deeper insights into the limitations of current AI models and inform the development of architectures better suited for social reasoning.

\newpage
\bibliography{iclr2025_conference}

\begin{thebibliography}{94}
\providecommand{\natexlab}[1]{#1}
\providecommand{\url}[1]{\texttt{#1}}
\expandafter\ifx\csname urlstyle\endcsname\relax
  \providecommand{\doi}[1]{doi: #1}\else
  \providecommand{\doi}{doi: \begingroup \urlstyle{rm}\Url}\fi

\bibitem[Achiam et~al.(2023)Achiam, Adler, Agarwal, Ahmad, Akkaya, Aleman, Almeida, Altenschmidt, Altman, Anadkat, et~al.]{achiam2023gpt}
Josh Achiam, Steven Adler, Sandhini Agarwal, Lama Ahmad, Ilge Akkaya, Florencia~Leoni Aleman, Diogo Almeida, Janko Altenschmidt, Sam Altman, Shyamal Anadkat, et~al.
\newblock Gpt-4 technical report.
\newblock \emph{arXiv preprint arXiv:2303.08774}, 2023.

\bibitem[Alayrac et~al.(2022)Alayrac, Donahue, Luc, Miech, Barr, Hasson, Lenc, Mensch, Millican, Reynolds, et~al.]{alayrac2022flamingo}
Jean-Baptiste Alayrac, Jeff Donahue, Pauline Luc, Antoine Miech, Iain Barr, Yana Hasson, Karel Lenc, Arthur Mensch, Katherine Millican, Malcolm Reynolds, et~al.
\newblock Flamingo: a visual language model for few-shot learning.
\newblock \emph{Advances in neural information processing systems}, 35:\penalty0 23716--23736, 2022.

\bibitem[Anscombe(1956)]{anscombe1956intention}
G.~E.~M. Anscombe.
\newblock \emph{Intention}.
\newblock Harvard University Press, 1956.

\bibitem[Apperly(2010)]{Apperly2010mindreaders}
Ian Apperly.
\newblock \emph{Mindreaders: The Cognitive Basis of "Theory of Mind"}.
\newblock Psychology Press, New York, NY, 2010.

\bibitem[Bai et~al.(2023{\natexlab{a}})Bai, Bai, Chu, Cui, Dang, Deng, Fan, Ge, Han, Huang, et~al.]{bai2023qwen}
Jinze Bai, Shuai Bai, Yunfei Chu, Zeyu Cui, Kai Dang, Xiaodong Deng, Yang Fan, Wenbin Ge, Yu~Han, Fei Huang, et~al.
\newblock Qwen technical report.
\newblock \emph{arXiv preprint arXiv:2309.16609}, 2023{\natexlab{a}}.

\bibitem[Bai et~al.(2023{\natexlab{b}})Bai, Bai, Yang, Wang, Tan, Wang, Lin, Zhou, and Zhou]{Qwen-VL}
Jinze Bai, Shuai Bai, Shusheng Yang, Shijie Wang, Sinan Tan, Peng Wang, Junyang Lin, Chang Zhou, and Jingren Zhou.
\newblock Qwen-vl: A versatile vision-language model for understanding, localization, text reading, and beyond.
\newblock \emph{arXiv preprint arXiv:2308.12966}, 2023{\natexlab{b}}.

\bibitem[Bai et~al.(2024)Bai, Du, Liang, Jin, Liu, Zhou, Zheng, Zhang, Ma, Wang, et~al.]{bai2024coig}
Yuelin Bai, Xinrun Du, Yiming Liang, Yonggang Jin, Ziqiang Liu, Junting Zhou, Tianyu Zheng, Xincheng Zhang, Nuo Ma, Zekun Wang, et~al.
\newblock Coig-cqia: Quality is all you need for chinese instruction fine-tuning.
\newblock \emph{arXiv preprint arXiv:2403.18058}, 2024.

\bibitem[Baker et~al.(2009)Baker, Saxe, and Tenenbaum]{baker2009action}
Chris~L Baker, Rebecca Saxe, and Joshua~B Tenenbaum.
\newblock Action understanding as inverse planning.
\newblock \emph{Cognition}, 113\penalty0 (3):\penalty0 329--349, 2009.

\bibitem[Barnes-Holmes et~al.(2004)Barnes-Holmes, McHugh, and Barnes-Holmes]{Barnes-Holmes2004perspective}
Yvonne Barnes-Holmes, Louise McHugh, and Dermot Barnes-Holmes.
\newblock Perspective-taking and theory of mind: A relational frame account.
\newblock \emph{The Behavior Analyst Today}, 5\penalty0 (1):\penalty0 15--25, 2004.

\bibitem[Bianco et~al.(2024)Bianco, Finisguerra, and Urgesi]{Bianco2024action}
Valentina Bianco, Alessandra Finisguerra, and Cosimo Urgesi.
\newblock Contextual priors shape action understanding before and beyond the unfolding of movement kinematics.
\newblock \emph{Brain Sciences}, 14\penalty0 (2):\penalty0 164, 2024.

\bibitem[Brass et~al.(2007)Brass, Schmitt, Spengler, and Gergely]{brass2007investigating}
Marcel Brass, Ruth~M Schmitt, Stephanie Spengler, and Gy{\"o}rgy Gergely.
\newblock Investigating action understanding: inferential processes versus action simulation.
\newblock \emph{Current biology}, 17\penalty0 (24):\penalty0 2117--2121, 2007.

\bibitem[Brown et~al.(2020)Brown, Mann, Ryder, Subbiah, Kaplan, Dhariwal, Neelakantan, Shyam, Sastry, Askell, et~al.]{brown2020language}
Tom Brown, Benjamin Mann, Nick Ryder, Melanie Subbiah, Jared~D Kaplan, Prafulla Dhariwal, Arvind Neelakantan, Pranav Shyam, Girish Sastry, Amanda Askell, et~al.
\newblock Language models are few-shot learners.
\newblock \emph{Advances in neural information processing systems}, 33:\penalty0 1877--1901, 2020.

\bibitem[Cai et~al.(2024)Cai, Ponomarenko, Yuan, Li, Yang, Dong, and Zhao]{cai2024spatialbot}
Wenxiao Cai, Yaroslav Ponomarenko, Jianhao Yuan, Xiaoqi Li, Wankou Yang, Hao Dong, and Bo~Zhao.
\newblock Spatialbot: Precise spatial understanding with vision language models.
\newblock \emph{arXiv preprint arXiv:2406.13642}, 2024.

\bibitem[Chen et~al.(2024)Chen, Xu, Kirmani, Ichter, Sadigh, Guibas, and Xia]{chen2024spatialvlm}
Boyuan Chen, Zhuo Xu, Sean Kirmani, Brain Ichter, Dorsa Sadigh, Leonidas Guibas, and Fei Xia.
\newblock Spatialvlm: Endowing vision-language models with spatial reasoning capabilities.
\newblock In \emph{Proceedings of the IEEE/CVF Conference on Computer Vision and Pattern Recognition}, pp.\  14455--14465, 2024.

\bibitem[Chen et~al.(2017)Chen, Papandreou, Kokkinos, Murphy, and Yuille]{chen2017deeplab}
Liang-Chieh Chen, George Papandreou, Iasonas Kokkinos, Kevin Murphy, and Alan~L Yuille.
\newblock Deeplab: Semantic image segmentation with deep convolutional nets, atrous convolution, and fully connected crfs.
\newblock \emph{IEEE transactions on pattern analysis and machine intelligence}, 40\penalty0 (4):\penalty0 834--848, 2017.

\bibitem[claude()]{claude}
claude.
\newblock Claude models - anthropic.
\newblock \url{https://docs.anthropic.com/en/docs/welcome#models}.

\bibitem[Cordts et~al.(2016)Cordts, Omran, Ramos, Rehfeld, Enzweiler, Benenson, Franke, Roth, and Schiele]{cordts2016cityscapes}
Marius Cordts, Mohamed Omran, Sebastian Ramos, Timo Rehfeld, Markus Enzweiler, Rodrigo Benenson, Uwe Franke, Stefan Roth, and Bernt Schiele.
\newblock The cityscapes dataset for semantic urban scene understanding.
\newblock In \emph{Proceedings of the IEEE conference on computer vision and pattern recognition}, pp.\  3213--3223, 2016.

\bibitem[Dai et~al.(2023)Dai, Li, Li, Tiong, Zhao, Wang, Li, Fung, and Hoi]{instructblip}
Wenliang Dai, Junnan Li, Dongxu Li, Anthony Meng~Huat Tiong, Junqi Zhao, Weisheng Wang, Boyang Li, Pascale Fung, and Steven Hoi.
\newblock Instructblip: Towards general-purpose vision-language models with instruction tuning, 2023.

\bibitem[Dennett(1987)]{dennett1987intentional}
Daniel~C. Dennett.
\newblock \emph{The Intentional Stance}.
\newblock MIT Press, Cambridge, MA, 1987.

\bibitem[Ford(1979)]{ford1979construct}
Martin~E Ford.
\newblock The construct validity of egocentrism.
\newblock \emph{Psychological Bulletin}, 86\penalty0 (6):\penalty0 1169, 1979.

\bibitem[Frith \& Frith(2005)Frith and Frith]{frith2005theory}
Chris Frith and Uta Frith.
\newblock Theory of mind.
\newblock \emph{Current biology}, 15\penalty0 (17):\penalty0 R644--R645, 2005.

\bibitem[Frith \& Frith(2006)Frith and Frith]{frith2006neural}
Chris~D Frith and Uta Frith.
\newblock The neural basis of mentalizing.
\newblock \emph{Neuron}, 50\penalty0 (4):\penalty0 531--534, 2006.

\bibitem[Fu et~al.(2023)Fu, Chen, Shen, Qin, Zhang, Lin, Yang, Zheng, Li, Sun, Wu, and Ji]{fu2023mme}
Chaoyou Fu, Peixian Chen, Yunhang Shen, Yulei Qin, Mengdan Zhang, Xu~Lin, Jinrui Yang, Xiawu Zheng, Ke~Li, Xing Sun, Yunsheng Wu, and Rongrong Ji.
\newblock Mme: A comprehensive evaluation benchmark for multimodal large language models.
\newblock \emph{arXiv preprint arXiv: 2306.13394}, 2023.

\bibitem[Gemini()]{gemini}
Gemini.
\newblock Gemini models| gemini api| google ai for developers.
\newblock \url{https://ai.google.dev/gemini-api/docs/models/gemini}.

\bibitem[Gemini(2023)]{team2023gemini}
Gemini.
\newblock Gemini: A family of highly capable multimodal models.
\newblock \emph{arXiv preprint arXiv: 2312.11805}, 2023.

\bibitem[Goddu et~al.(2024)Goddu, No{\"e}, and Thompson]{goddu2024llms}
Mariel~K Goddu, Alva No{\"e}, and Evan Thompson.
\newblock Llms don’t know anything: reply to yildirim and paul.
\newblock \emph{Trends in Cognitive Sciences}, 2024.

\bibitem[Gomez-Marin et~al.(2014)Gomez-Marin, Paton, Kampff, Costa, and Mainen]{gomez2014big}
Alex Gomez-Marin, Joseph~J Paton, Adam~R Kampff, Rui~M Costa, and Zachary~F Mainen.
\newblock Big behavioral data: psychology, ethology and the foundations of neuroscience.
\newblock \emph{Nature neuroscience}, 17\penalty0 (11):\penalty0 1455--1462, 2014.

\bibitem[Gopnik \& Wellman(1992)Gopnik and Wellman]{gopnik1992child}
Alison Gopnik and Henry~M Wellman.
\newblock Why the child's theory of mind really is a theory.
\newblock \emph{Mind \& Language}, 7\penalty0 (1-2):\penalty0 145--171, 1992.

\bibitem[Gunia et~al.(2021)Gunia, Moraresku, and Vl{\v{c}}ek]{gunia2021brain}
Anna Gunia, Sofiia Moraresku, and Kamil Vl{\v{c}}ek.
\newblock Brain mechanisms of visuospatial perspective-taking in relation to object mental rotation and the theory of mind.
\newblock \emph{Behavioural Brain Research}, 407:\penalty0 113247, 2021.

\bibitem[Gweon(2021)]{gweon2021inferential}
Hyowon Gweon.
\newblock Inferential social learning: Cognitive foundations of human social learning and teaching.
\newblock \emph{Trends in cognitive sciences}, 25\penalty0 (10):\penalty0 896--910, 2021.

\bibitem[Gweon \& Schulz(2011)Gweon and Schulz]{gweon201116}
Hyowon Gweon and Laura Schulz.
\newblock 16-month-olds rationally infer causes of failed actions.
\newblock \emph{Science}, 332\penalty0 (6037):\penalty0 1524--1524, 2011.

\bibitem[Hao et~al.(2023)Hao, Gu, Ma, Hong, Wang, Wang, and Hu]{hao2023reasoning}
Shibo Hao, Yi~Gu, Haodi Ma, Joshua~Jiahua Hong, Zhen Wang, Daisy~Zhe Wang, and Zhiting Hu.
\newblock Reasoning with language model is planning with world model.
\newblock \emph{arXiv preprint arXiv:2305.14992}, 2023.

\bibitem[Herath et~al.(2017)Herath, Harandi, and Porikli]{herath2017going}
Samitha Herath, Mehrtash Harandi, and Fatih Porikli.
\newblock Going deeper into action recognition: A survey.
\newblock \emph{Image and vision computing}, 60:\penalty0 4--21, 2017.

\bibitem[Hong et~al.(2024)Hong, Wang, Ding, Yu, Lv, Wang, Cheng, Huang, Ji, Xue, et~al.]{hong2024cogvlm2}
Wenyi Hong, Weihan Wang, Ming Ding, Wenmeng Yu, Qingsong Lv, Yan Wang, Yean Cheng, Shiyu Huang, Junhui Ji, Zhao Xue, et~al.
\newblock Cogvlm2: Visual language models for image and video understanding.
\newblock \emph{arXiv preprint arXiv:2408.16500}, 2024.

\bibitem[Jaech et~al.(2024)Jaech, Kalai, Lerer, Richardson, El-Kishky, Low, Helyar, Madry, Beutel, Carney, et~al.]{jaech2024openai}
Aaron Jaech, Adam Kalai, Adam Lerer, Adam Richardson, Ahmed El-Kishky, Aiden Low, Alec Helyar, Aleksander Madry, Alex Beutel, Alex Carney, et~al.
\newblock Openai o1 system card.
\newblock \emph{arXiv preprint arXiv:2412.16720}, 2024.

\bibitem[Jhuang et~al.(2013)Jhuang, Gall, Zuffi, Schmid, and Black]{jhuang2013towards}
Hueihan Jhuang, Juergen Gall, Silvia Zuffi, Cordelia Schmid, and Michael~J Black.
\newblock Towards understanding action recognition.
\newblock In \emph{Proceedings of the IEEE international conference on computer vision}, pp.\  3192--3199, 2013.

\bibitem[Jiang et~al.(2023)Jiang, Sablayrolles, Mensch, Bamford, Chaplot, de~las Casas, Bressand, Lengyel, Lample, Saulnier, Lavaud, Lachaux, Stock, Scao, Lavril, Wang, Lacroix, and Sayed]{jiang2023mistral}
Albert~Q. Jiang, Alexandre Sablayrolles, Arthur Mensch, Chris Bamford, Devendra~Singh Chaplot, Diego de~las Casas, Florian Bressand, Gianna Lengyel, Guillaume Lample, Lucile Saulnier, Lélio~Renard Lavaud, Marie-Anne Lachaux, Pierre Stock, Teven~Le Scao, Thibaut Lavril, Thomas Wang, Timothée Lacroix, and William~El Sayed.
\newblock Mistral 7b.
\newblock \emph{arXiv preprint arXiv: 2310.06825}, 2023.

\bibitem[Johnson(1975)]{johnson1975cooperativeness}
David~W Johnson.
\newblock Cooperativeness and social perspective taking.
\newblock \emph{Journal of Personality and Social Psychology}, 31\penalty0 (2):\penalty0 241, 1975.

\bibitem[Johnson-Laird(1983)]{johnson1983mental}
Philip~Nicholas Johnson-Laird.
\newblock \emph{Mental models: Towards a cognitive science of language, inference, and consciousness}.
\newblock Number~6. Harvard University Press, 1983.

\bibitem[Kaplan et~al.(2020)Kaplan, McCandlish, Henighan, Brown, Chess, Child, Gray, Radford, Wu, and Amodei]{kaplan2020scaling}
Jared Kaplan, Sam McCandlish, Tom Henighan, Tom~B Brown, Benjamin Chess, Rewon Child, Scott Gray, Alec Radford, Jeffrey Wu, and Dario Amodei.
\newblock Scaling laws for neural language models.
\newblock \emph{arXiv preprint arXiv:2001.08361}, 2020.

\bibitem[Kilner(2004)]{Kilner2011action}
James~M Kilner.
\newblock More than one pathway to action understanding.
\newblock \emph{Trends in cognitive sciences}, 15\penalty0 (8):\penalty0 352--357, 2004.

\bibitem[Kong \& Fu(2022)Kong and Fu]{kong2022human}
Yu~Kong and Yun Fu.
\newblock Human action recognition and prediction: A survey.
\newblock \emph{International Journal of Computer Vision}, 130\penalty0 (5):\penalty0 1366--1401, 2022.

\bibitem[Kosinski(2023)]{Kosinski2023ToM}
Michal Kosinski.
\newblock Evaluating large language models in theory of mind tasksg.
\newblock \emph{arXiv preprint arXiv:2302.02083}, 2023.

\bibitem[Lamm et~al.(2007)Lamm, Batson, and Decety]{lamm2007neural}
Claus Lamm, C~Daniel Batson, and Jean Decety.
\newblock The neural substrate of human empathy: effects of perspective-taking and cognitive appraisal.
\newblock \emph{Journal of cognitive neuroscience}, 19\penalty0 (1):\penalty0 42--58, 2007.

\bibitem[Lan et~al.(2014)Lan, Chen, and Savarese]{lan2014hierarchical}
Tian Lan, Tsung-Chuan Chen, and Silvio Savarese.
\newblock A hierarchical representation for future action prediction.
\newblock In \emph{Computer Vision--ECCV 2014: 13th European Conference, Zurich, Switzerland, September 6-12, 2014, Proceedings, Part III 13}, pp.\  689--704. Springer, 2014.

\bibitem[LeCun(2022)]{lecun2022path}
Yann LeCun.
\newblock A path towards autonomous machine intelligence version 0.9. 2, 2022-06-27.
\newblock \emph{Open Review}, 62\penalty0 (1):\penalty0 1--62, 2022.

\bibitem[Lehmann \& Jansen(2019)Lehmann and Jansen]{lehmann2019relationship}
Jennifer Lehmann and Petra Jansen.
\newblock The relationship between theory of mind and mental rotation ability in preschool-aged children.
\newblock \emph{Cogent Psychology}, 6\penalty0 (1):\penalty0 1582127, 2019.

\bibitem[Li et~al.(2024{\natexlab{a}})Li, Ge, Ge, Wang, Wang, Zhang, and Shan]{li2024seed}
Bohao Li, Yuying Ge, Yixiao Ge, Guangzhi Wang, Rui Wang, Ruimao Zhang, and Ying Shan.
\newblock Seed-bench: Benchmarking multimodal large language models.
\newblock In \emph{Proceedings of the IEEE/CVF Conference on Computer Vision and Pattern Recognition}, pp.\  13299--13308, 2024{\natexlab{a}}.

\bibitem[Li et~al.(2023{\natexlab{a}})Li, Li, Savarese, and Hoi]{li2023blip}
Junnan Li, Dongxu Li, Silvio Savarese, and Steven Hoi.
\newblock Blip-2: Bootstrapping language-image pre-training with frozen image encoders and large language models.
\newblock In \emph{International conference on machine learning}, pp.\  19730--19742. PMLR, 2023{\natexlab{a}}.

\bibitem[Li et~al.(2023{\natexlab{b}})Li, Li, Savarese, and Hoi]{li2023blip2}
Junnan Li, Dongxu Li, Silvio Savarese, and Steven Hoi.
\newblock {BLIP-2:} bootstrapping language-image pre-training with frozen image encoders and large language models.
\newblock In \emph{ICML}, 2023{\natexlab{b}}.

\bibitem[Li et~al.(2019)Li, Yatskar, Yin, Hsieh, and Chang]{li2019visualbert}
Liunian~Harold Li, Mark Yatskar, Da~Yin, Cho-Jui Hsieh, and Kai-Wei Chang.
\newblock Visualbert: A simple and performant baseline for vision and language.
\newblock \emph{arXiv preprint arXiv:1908.03557}, 2019.

\bibitem[Li et~al.(2022)Li, Cai, Gao, Li, and Hu]{li2022more}
Yijiang Li, Wentian Cai, Ying Gao, Chengming Li, and Xiping Hu.
\newblock More than encoder: Introducing transformer decoder to upsample.
\newblock In \emph{2022 IEEE international conference on bioinformatics and biomedicine (BIBM)}, pp.\  1597--1602. IEEE, 2022.

\bibitem[Li et~al.(2023{\natexlab{c}})Li, Wang, Yang, Feng, Zhang, and Gao]{li2023diverse}
Yijiang Li, Xinjiang Wang, Lihe Yang, Litong Feng, Wayne Zhang, and Ying Gao.
\newblock Diverse cotraining makes strong semi-supervised segmentor.
\newblock \emph{arXiv preprint arXiv:2308.09281}, 2023{\natexlab{c}}.

\bibitem[Li et~al.(2024{\natexlab{b}})Li, Gao, Zhao, Wang, Sun, Lyu, Hawkins, Vasconcelos, Golan, Luo, et~al.]{li2024core}
Yijiang Li, Qingying Gao, Tianwei Zhao, Bingyang Wang, Haoran Sun, Haiyun Lyu, Robert~D Hawkins, Nuno Vasconcelos, Tal Golan, Dezhi Luo, et~al.
\newblock Core knowledge deficits in multi-modal language models.
\newblock \emph{arXiv preprint arXiv:2410.10855}, 2024{\natexlab{b}}.

\bibitem[Li et~al.(2025)Li, Zhang, Cheng, Li, Shan, Gao, Lyu, Li, Bi, and Vasconcelos]{li2025egoprivacy}
Yijiang Li, Genpei Zhang, Jiacheng Cheng, Yi~Li, Xiaojun Shan, Dashan Gao, Jiancheng Lyu, Yuan Li, Ning Bi, and Nuno Vasconcelos.
\newblock Egoprivacy: What your first-person camera says about you?
\newblock \emph{arXiv preprint arXiv:2506.12258}, 2025.

\bibitem[Liu et~al.(2023{\natexlab{a}})Liu, Li, Li, and Lee]{liu2023improvedllava}
Haotian Liu, Chunyuan Li, Yuheng Li, and Yong~Jae Lee.
\newblock Improved baselines with visual instruction tuning, 2023{\natexlab{a}}.

\bibitem[Liu et~al.(2023{\natexlab{b}})Liu, Li, Wu, and Lee]{liu2023llava}
Haotian Liu, Chunyuan Li, Qingyang Wu, and Yong~Jae Lee.
\newblock Visual instruction tuning, 2023{\natexlab{b}}.

\bibitem[Liu et~al.(2024{\natexlab{a}})Liu, Li, Li, Li, Zhang, Shen, and Lee]{liu2024llavanext}
Haotian Liu, Chunyuan Li, Yuheng Li, Bo~Li, Yuanhan Zhang, Sheng Shen, and Yong~Jae Lee.
\newblock Llava-next: Improved reasoning, ocr, and world knowledge, January 2024{\natexlab{a}}.
\newblock URL \url{https://llava-vl.github.io/blog/2024-01-30-llava-next/}.

\bibitem[Liu et~al.(2024{\natexlab{b}})Liu, Li, Wu, and Lee]{liu2024visual}
Haotian Liu, Chunyuan Li, Qingyang Wu, and Yong~Jae Lee.
\newblock Visual instruction tuning.
\newblock \emph{Advances in neural information processing systems}, 36, 2024{\natexlab{b}}.

\bibitem[Liu et~al.(2017)Liu, Ullman, Tenenbaum, and Spelke]{liu2017ten}
Shari Liu, Tomer~D Ullman, Joshua~B Tenenbaum, and Elizabeth~S Spelke.
\newblock Ten-month-old infants infer the value of goals from the costs of actions.
\newblock \emph{Science}, 358\penalty0 (6366):\penalty0 1038--1041, 2017.

\bibitem[Mitchell \& Krakauer(2023)Mitchell and Krakauer]{mitchell2023debate}
Melanie Mitchell and David~C Krakauer.
\newblock The debate over understanding in ai’s large language models.
\newblock \emph{Proceedings of the National Academy of Sciences}, 120\penalty0 (13):\penalty0 e2215907120, 2023.

\bibitem[Moll \& Meltzoff(2011)Moll and Meltzoff]{Moll2011look}
Henrike Moll and Andrew~N Meltzoff.
\newblock How does it look? level 2 perspective-taking at 36 months of age.
\newblock \emph{Child Development}, 82\penalty0 (2):\penalty0 661--673, 2011.

\bibitem[Mori et~al.(1999)Mori, Nishida, and Yamada]{mori1999optical}
Shunji Mori, Hirobumi Nishida, and Hiromitsu Yamada.
\newblock \emph{Optical character recognition}.
\newblock John Wiley \& Sons, Inc., 1999.

\bibitem[OpenAI()]{gpt4o}
OpenAI.
\newblock Models - openai api.
\newblock \url{https://platform.openai.com/docs/models/gpt-4o}.

\bibitem[OpenAI(2023)]{openai2023gpt4}
OpenAI.
\newblock Gpt-4 technical report.
\newblock \emph{arXiv preprint arXiv: 2303.08774}, 2023.

\bibitem[Peng et~al.(2023)Peng, Li, He, Galley, and Gao]{peng2023instruction}
Baolin Peng, Chunyuan Li, Pengcheng He, Michel Galley, and Jianfeng Gao.
\newblock Instruction tuning with gpt-4.
\newblock \emph{arXiv preprint arXiv:2304.03277}, 2023.

\bibitem[Piaget(1977)]{piaget1977development}
Jean Piaget.
\newblock \emph{The Development of Thought: Equilibration of Cognitive Structures}.
\newblock Viking Press, 1977.

\bibitem[Piaget \& Inhelder(1957)Piaget and Inhelder]{piaget1956conception}
Jean Piaget and B{\"a}rbel Inhelder.
\newblock \emph{The Child's Conception of Space}.
\newblock Routledge, London, 1957.

\bibitem[Piaget \& Inhelder(1969)Piaget and Inhelder]{piaget1969psychology}
Jean Piaget and B{\"a}rbel Inhelder.
\newblock \emph{The Psychology of the Child}.
\newblock Basic Books, New York, 1969.

\bibitem[Premack \& Woodruff(1978)Premack and Woodruff]{premack1978chimpanzee}
David~G. Premack and Guy Woodruff.
\newblock Does the chimpanzee have a theory of mind?
\newblock \emph{Behavioral and Brain Sciences}, 1978.

\bibitem[Radford et~al.(2021)Radford, Kim, Hallacy, Ramesh, Goh, Agarwal, Sastry, Askell, Mishkin, Clark, Krueger, and Sutskever]{radford2021learning}
Alec Radford, Jong~Wook Kim, Chris Hallacy, Aditya Ramesh, Gabriel Goh, Sandhini Agarwal, Girish Sastry, Amanda Askell, Pamela Mishkin, Jack Clark, Gretchen Krueger, and Ilya Sutskever.
\newblock Learning transferable visual models from natural language supervision.
\newblock \emph{arXiv preprint arXiv: 2103.00020}, 2021.

\bibitem[Rahmanzadehgervi et~al.(2024)Rahmanzadehgervi, Bolton, Taesiri, and Nguyen]{rahmanzadehgervi2024vision}
Pooyan Rahmanzadehgervi, Logan Bolton, Mohammad~Reza Taesiri, and Anh~Totti Nguyen.
\newblock Vision language models are blind.
\newblock In \emph{Proceedings of the Asian Conference on Computer Vision}, pp.\  18--34, 2024.

\bibitem[Rosenthal(1991)]{rosenthal1991nature}
David~M. Rosenthal.
\newblock \emph{The Nature of Mind}.
\newblock Oxford University Press, New York, 1991.

\bibitem[Searle(1979)]{searle1979intentionality}
John~R Searle.
\newblock The intentionality of intention and action.
\newblock \emph{Inquiry}, 22\penalty0 (1-4):\penalty0 253--280, 1979.

\bibitem[Searle(1980)]{searle1980minds}
John~R Searle.
\newblock Minds, brains and programs.
\newblock \emph{Behavioral and Brain Sciences}, 1980.

\bibitem[Shanton \& Goldman(2010)Shanton and Goldman]{shanton2010simulation}
Karen Shanton and Alvin Goldman.
\newblock Simulation theory.
\newblock \emph{Wiley Interdisciplinary Reviews: Cognitive Science}, 1\penalty0 (4):\penalty0 527--538, 2010.

\bibitem[Shao et~al.(2024)Shao, Qian, Xiao, Song, Zong, Wang, Liu, and Li]{shao2024visual}
Hao Shao, Shengju Qian, Han Xiao, Guanglu Song, Zhuofan Zong, Letian Wang, Yu~Liu, and Hongsheng Li.
\newblock Visual cot: Unleashing chain-of-thought reasoning in multi-modal language models.
\newblock \emph{arXiv preprint arXiv:2403.16999}, 2024.

\bibitem[Shi et~al.(2024)Shi, Ye, Fang, Jin, Isik, Kuo, and Shu]{shi2024muma}
Haojun Shi, Suyu Ye, Xinyu Fang, Chuanyang Jin, Leyla Isik, Yen-Ling Kuo, and Tianmin Shu.
\newblock Muma-tom: Multi-modal multi-agent theory of mind.
\newblock \emph{arXiv preprint arXiv:2408.12574}, 2024.

\bibitem[Shu et~al.(2021)Shu, Bhandwaldar, Gan, Smith, Liu, Gutfreund, Spelke, Tenenbaum, and Ullman]{shu2021agent}
Tianmin Shu, Abhishek Bhandwaldar, Chuang Gan, Kevin Smith, Shari Liu, Dan Gutfreund, Elizabeth Spelke, Joshua Tenenbaum, and Tomer Ullman.
\newblock Agent: A benchmark for core psychological reasoning.
\newblock In \emph{International conference on machine learning}, pp.\  9614--9625. PMLR, 2021.

\bibitem[Strachan et~al.(2024)Strachan, Albergo, Borghini, Pansardi, Scaliti, Gupta, Saxena, Rufo, Panzeri, Manzi, Graziano, and Becchio]{Strachan2024ToM}
James W~A Strachan, Dalila Albergo, Giulia Borghini, Oriana Pansardi, Eugenio Scaliti, Saurabh Gupta, Krati Saxena, Alessandro Rufo, Stefano Panzeri, Guido Manzi, Michael S~A Graziano, and Cristina Becchio.
\newblock Testing theory of mind in large language models and humans.
\newblock \emph{Nature Human Behaviour}, 8\penalty0 (7):\penalty0 1285--1295, 2024.

\bibitem[Sutton(2019)]{sutton2019bitter}
Richard Sutton.
\newblock The bitter lesson.
\newblock \emph{Incomplete Ideas (blog)}, 13\penalty0 (1):\penalty0 38, 2019.

\bibitem[Tan \& Bansal(2019)Tan and Bansal]{tan2019lxmert}
Hao Tan and Mohit Bansal.
\newblock Lxmert: Learning cross-modality encoder representations from transformers.
\newblock \emph{arXiv preprint arXiv:1908.07490}, 2019.

\bibitem[Touvron et~al.(2023)Touvron, Lavril, Izacard, Martinet, Lachaux, Lacroix, Rozi{\`e}re, Goyal, Hambro, Azhar, et~al.]{touvron2023llama}
Hugo Touvron, Thibaut Lavril, Gautier Izacard, Xavier Martinet, Marie-Anne Lachaux, Timoth{\'e}e Lacroix, Baptiste Rozi{\`e}re, Naman Goyal, Eric Hambro, Faisal Azhar, et~al.
\newblock Llama: Open and efficient foundation language models.
\newblock \emph{arXiv preprint arXiv:2302.13971}, 2023.

\bibitem[Van~Rooij et~al.(2024)Van~Rooij, Guest, Adolfi, de~Haan, Kolokolova, and Rich]{van2024reclaiming}
Iris Van~Rooij, Olivia Guest, Federico Adolfi, Ronald de~Haan, Antonina Kolokolova, and Patricia Rich.
\newblock Reclaiming ai as a theoretical tool for cognitive science.
\newblock \emph{Computational Brain \& Behavior}, 7\penalty0 (4):\penalty0 616--636, 2024.

\bibitem[Wang et~al.(2024)Wang, Bai, Tan, Wang, Fan, Bai, Chen, Liu, Wang, Ge, Fan, Dang, Du, Ren, Men, Liu, Zhou, Zhou, and Lin]{Qwen2VL}
Peng Wang, Shuai Bai, Sinan Tan, Shijie Wang, Zhihao Fan, Jinze Bai, Keqin Chen, Xuejing Liu, Jialin Wang, Wenbin Ge, Yang Fan, Kai Dang, Mengfei Du, Xuancheng Ren, Rui Men, Dayiheng Liu, Chang Zhou, Jingren Zhou, and Junyang Lin.
\newblock Qwen2-vl: Enhancing vision-language model's perception of the world at any resolution.
\newblock \emph{arXiv preprint arXiv:2409.12191}, 2024.

\bibitem[Wang et~al.(2023)Wang, Yang, Zhang, Li, Feng, Fang, Lyu, Chen, and Zhang]{wang2023consistent}
Xinjiang Wang, Xingyi Yang, Shilong Zhang, Yijiang Li, Litong Feng, Shijie Fang, Chengqi Lyu, Kai Chen, and Wayne Zhang.
\newblock Consistent-teacher: Towards reducing inconsistent pseudo-targets in semi-supervised object detection.
\newblock In \emph{Proceedings of the IEEE/CVF conference on computer vision and pattern recognition}, pp.\  3240--3249, 2023.

\bibitem[Wellman(1992)]{wellman1992child}
Henry~M. Wellman.
\newblock \emph{The Child's Theory of Mind}.
\newblock MIT Press, Cambridge, MA, 1992.

\bibitem[Wu \& Xie(2024)Wu and Xie]{wu2024v}
Penghao Wu and Saining Xie.
\newblock V?: Guided visual search as a core mechanism in multimodal llms.
\newblock In \emph{Proceedings of the IEEE/CVF Conference on Computer Vision and Pattern Recognition}, pp.\  13084--13094, 2024.

\bibitem[Xu et~al.(2024)Xu, Jin, Hao, Song, Sun, and Yuan]{xu2024llava}
Guowei Xu, Peng Jin, Li~Hao, Yibing Song, Lichao Sun, and Li~Yuan.
\newblock Llava-o1: Let vision language models reason step-by-step.
\newblock \emph{arXiv preprint arXiv:2411.10440}, 2024.

\bibitem[Xu et~al.(2023)Xu, Wu, Rosenman, Lal, Che, and Duan]{xu2023bridgetower}
Xiao Xu, Chenfei Wu, Shachar Rosenman, Vasudev Lal, Wanxiang Che, and Nan Duan.
\newblock Bridgetower: Building bridges between encoders in vision-language representation learning.
\newblock In \emph{Proceedings of the AAAI Conference on Artificial Intelligence}, volume~37, pp.\  10637--10647, 2023.

\bibitem[Yildirim \& Paul(2024)Yildirim and Paul]{yildirim2024task}
Ilker Yildirim and LA~Paul.
\newblock From task structures to world models: what do llms know?
\newblock \emph{Trends in Cognitive Sciences}, 2024.

\bibitem[Zhang et~al.(2024)Zhang, Xie, Feng, Li, Xing, Zheng, and Lu]{zhang2024pixels}
Wanpeng Zhang, Zilong Xie, Yicheng Feng, Yijiang Li, Xingrun Xing, Sipeng Zheng, and Zongqing Lu.
\newblock From pixels to tokens: Byte-pair encoding on quantized visual modalities.
\newblock \emph{arXiv preprint arXiv:2410.02155}, 2024.

\bibitem[Zhang et~al.(2025)Zhang, Feng, Luo, Li, Yue, Zheng, and Lu]{zhang2025unified}
Wanpeng Zhang, Yicheng Feng, Hao Luo, Yijiang Li, Zihao Yue, Sipeng Zheng, and Zongqing Lu.
\newblock Unified multimodal understanding via byte-pair visual encoding.
\newblock \emph{arXiv preprint arXiv:2506.23639}, 2025.

\bibitem[Zhao et~al.(2016)Zhao, Malle, and Gweon]{zhao2016nine}
Xuan Zhao, Bertram~F Malle, and Hyowon Gweon.
\newblock Is it a nine, or a six? prosocial and selective perspective taking in four-year-olds.
\newblock In \emph{Proceedings of the Annual Meeting of the Cognitive Science Society}, volume~38, pp.\  924--929, 2016.

\end{thebibliography}
\bibliographystyle{iclr2025_conference}

\newpage
\appendix

\section{List of Models Evaluated}

\begin{table}[h!]
\centering
\scriptsize
\begin{tabular}{llll}
\toprule
Models \\
\midrule
GPT4V\_TURBO\_HIGH & GPT4o\_HIGH & GPT4o\_MINI & InternVL-Chat-V1-1 \\
InternVL2-1B & InternVL2-2B & InternVL2-4B & Mantis-8B-clip-llama3 \\
Mantis-8B-siglip-llama3 & Mini-InternVL-Chat-2B-V1-5 & Mini-InternVL-Chat-4B-V1-5 & MiniCPM-V \\
MiniCPM-V-2 & OmChat & Ovis1.5-Llama3-8B & Phi-3-Vision \\
Phi-3.5-Vision & Pixtral-12B & Qwen2-VL-2B-Instruct & Qwen2-VL-7B-Instruct \\
Slime-13B & Slime-7B & Slime-8B & deepseek\_vl\_1.3b \\
deepseek\_vl\_7b & idefics2\_8b & llava\_next\_interleave\_7b & llava\_next\_interleave\_7b\_dpo \\
llava\_next\_llama3 & llava\_next\_mistral\_7b & llava\_onevision\_qwen2\_0.5b\_si & llava\_onevision\_qwen2\_7b\_ov \\
llava\_v1.5\_13b & llava\_v1.5\_7b & mPLUG-Owl3 & qwen\_chat \\
sharegpt4v\_7b & & & \\
\bottomrule
\end{tabular}
\caption{List of Models}
\label{tab:models}
\end{table}

\newpage

\section{Additional Examples}

\begin{figure}[h]
  \centering
  \includegraphics{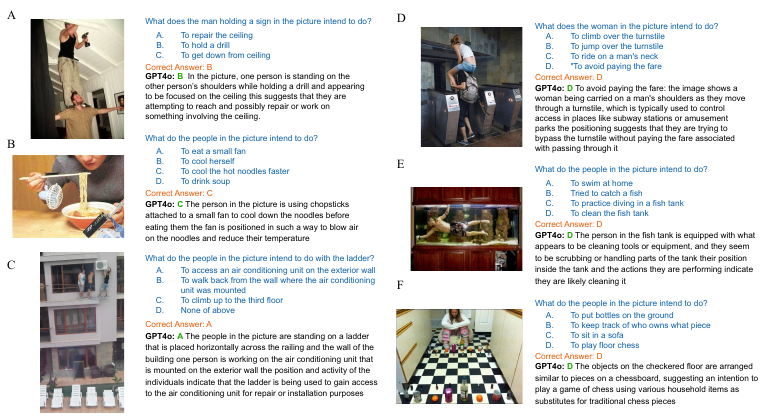}
  \caption{Additional Examples of Vision Language Models Assessed with IntentBench: Correct}.
  \label{fig:fig4}
\end{figure}

\begin{figure}[h]
  \centering
  \includegraphics{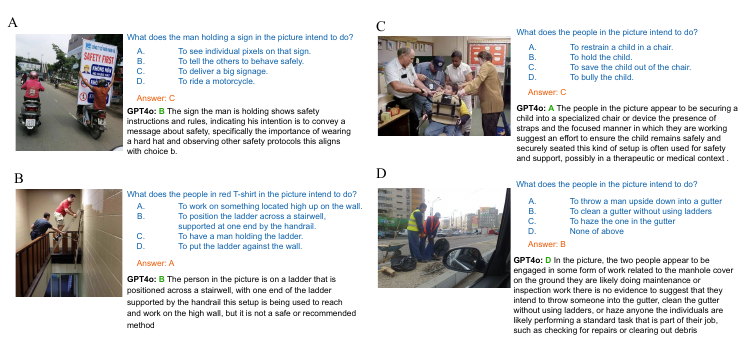}
  \caption{Additional Examples of Vision Language Models Assessed with IntentBench: Wrong}.
  \label{fig:fig7}
\end{figure}



\end{document}